\documentclass[10pt,twocolumn,letterpaper]{article}

\usepackage{cvpr}              

\usepackage[dvipsnames]{xcolor}
\usepackage{microtype}
\usepackage{xspace}
\usepackage{booktabs}
\usepackage{multirow}
\usepackage{tabularx}
\usepackage{comment}

\newcommand{\SN}{SpaceNet~8\xspace}
\newcommand{\ours}{ODEED\xspace}

\renewcommand{\paragraph}[1]{\medskip\noindent{\textbf{#1}}}

\definecolor{cvprblue}{rgb}{0.21,0.49,0.74}
\usepackage[pagebackref,breaklinks,colorlinks,citecolor=cvprblue]{hyperref}
\usepackage[capitalize]{cleveref}
\usepackage[accsupp]{axessibility}

\def\paperID{27} 
\def\confName{EarthVision}
\def\confYear{2024}

\title{Detecting Out-Of-Distribution Earth Observation Images with Diffusion Models}

\author{Georges Le Bellier \textsuperscript{1}\\
\textsuperscript{1} Cnam, CEDRIC, EA4629\\
F-75141 Paris, France\\
{\tt\small georges.le-bellier@lecnam.net}
\and
Nicolas Audebert \textsuperscript{1,2}\\
\textsuperscript{2} Univ. Gustave Eiffel, ENSG, IGN, LASTIG\\ F-94160 Saint-Mandé, France\\
{\tt\small nicolas.audebert@ign.fr}
}

\begin{document}
\maketitle
\begin{abstract}
Earth Observation imagery can capture rare and unusual events, such as disasters and major landscape changes, whose visual appearance contrasts with the usual observations. Deep models trained on common remote sensing data will output drastically different features for these out-of-distribution samples, compared to those closer to their training dataset.
Detecting them could therefore help anticipate changes in the observations, either geographical or environmental.
In this work, we show that the reconstruction error of diffusion models can effectively serve as unsupervised out-of-distribution detectors for remote sensing images, using them as a plausibility score.
Moreover, we introduce \ours, a novel reconstruction-based scorer using the probability-flow ODE of diffusion models. We validate it experimentally on \SN with various scenarios, such as classical OOD detection with geographical shift and near-OOD setups: pre/post-flood and non-flooded/flooded image recognition.
We show that our \ours scorer significantly outperforms other diffusion-based and discriminative baselines on the more challenging near-OOD scenarios of flood image detection, where OOD images are close to the distribution tail.
We aim to pave the way towards better use of generative models for anomaly detection in remote sensing.

\end{abstract}

\section{Introduction}
\label{sec:intro}

\begin{figure}[t]
    \begin{minipage}{0.5\textwidth}
        \centering
        \includegraphics[width=1\linewidth]{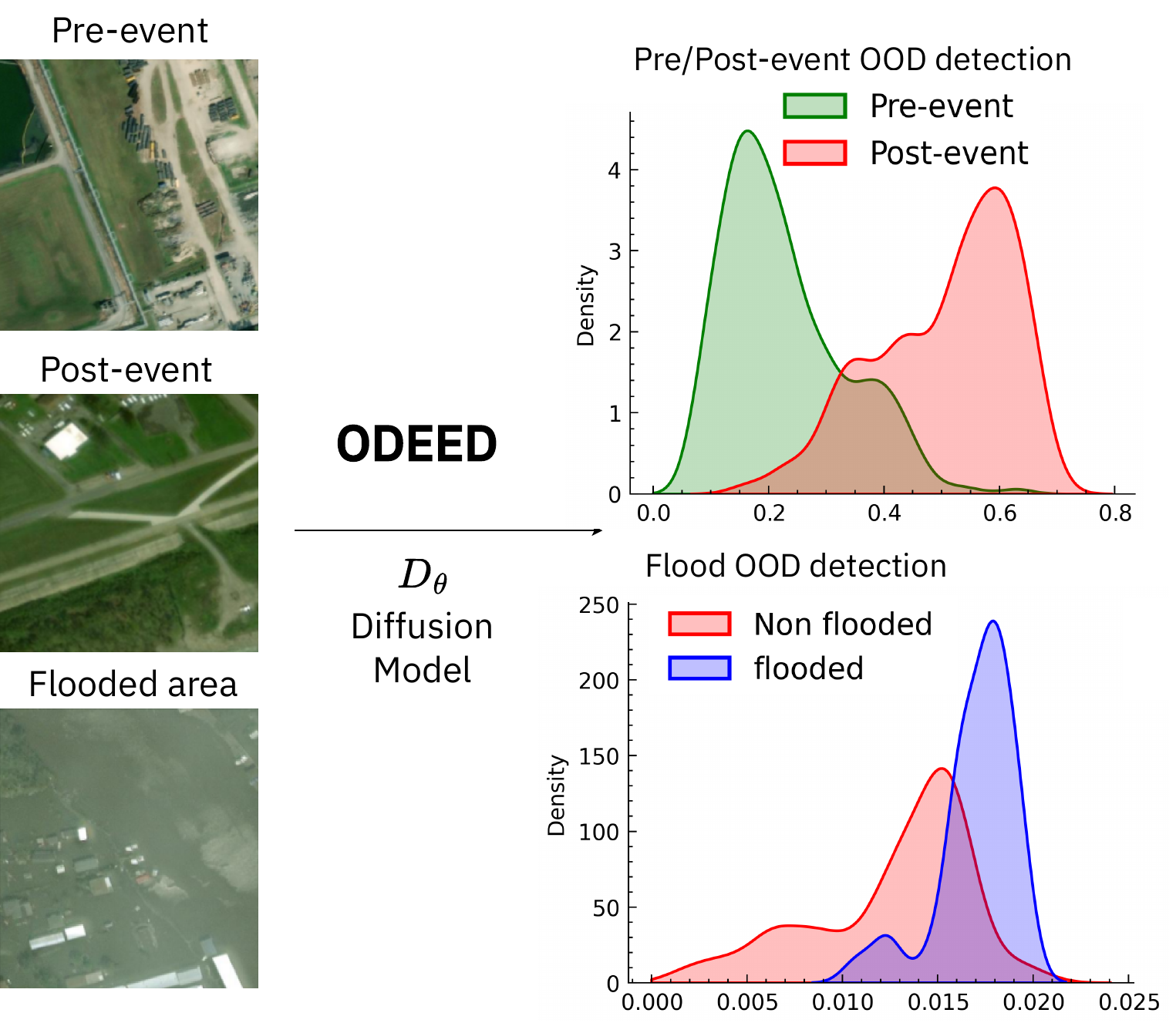}
        \caption{\ours discriminates between pre and post-event images by making larger reconstruction errors on the latter \wrt the LPIPS metric (top). It can also isolate flooded areas from the other post-event images when evaluating the reconstruction similarity with the MSE (bottom).}
        \label{fig:intro}
    \end{minipage}
\end{figure}

In recent years, deep learning has grown to be a staple of image understanding in computer vision, Earth Observation included.
Deep neural networks have been for several years now the state of the art for many tasks, from biomass estimation to land cover segmentation \cite{zhu_deep_2017}.
Despite their impressive capacity for generalization, these models are still trained on finite datasets. 
For predictive models, it is critical to be able to detect when new observations fall outside of this training set, as model performance can drop sharply.
This task called \textit{Out-Of-Distribution} (OOD) detection \cite{yang_generalized_2024a}, is a specific case of anomaly detection and represents a major challenge in improving the robustness of deep models.

Indeed, remote sensing imagery is facing a multitude of possible distribution shifts, from a change of acquisition sensor to a change of season to a change of geographical area. Being able to detect these OOD images is helpful to avoid considering degraded model predictions. In addition, disasters such as floods, forest fires, and storms are also unfrequent and catastrophic events, that are rarely observed in remote sensing datasets. While technically not always OOD, these events are ``near'' out-of-distribution observations as shown in \cref{fig:intro}, whose detection can be of great help for disaster management and surveillance. Identifying unusual observations can also help automate the curation of new datasets by eliminating outliers caused by cloud cover, sensor faults, or other artifacts.

Moreover, annotations are scarce in Earth Observation. The huge mass of available images remains largely unlabeled, due to the expert knowledge and time required to do so. For this reason, OOD detectors tailored to remote sensing data should be fully unsupervised, and be able to model the data distribution without labels. 
Generative models have been advanced as a way to model the distribution of the data and build likelihood estimators to detect OOD samples \cite{zhang_understanding_2021}.
In particular, recent works in classical computer vision have established \textit{diffusion models} as the new state-of-the-art to detect out-of-distribution images \cite{goodier2023likelihood}.
Diffusion models \cite{song2020generative} are a family of unsupervised generative models that model the underlying distribution of training examples. They have outclassed previous models in image generation, such as Generative Adversarial Networks, due to their strong ability to model complex distributions without mode collapse.
Promising new works have shown that diffusion models were applicable to Earth Observation imagery, \eg for cloud removal \cite{Sanguigni2023DiffusionMF} or image synthesis \cite{khanna2024diffusionsat}.

In this paper, we demonstrate the relevance of unconditional diffusion models for OOD detection in remote sensing. Several use cases are considered in our work: cloud detection, different geographical domains, pre- and post-disaster, and flooded/non-flooded images (cf. \cref{fig:intro}). We introduce a new \ours specifically tailored to the latter use cases and show it outperforms all existing baselines.

\section{Related Work}
\label{sec:related_work}

\subsection{Diffusion models}

Diffusion models are a powerful family of generative models that take inspiration from \textit{score-matching} \cite{JMLR:v6:hyvarinen05a, vincent, song2019generative}. They were first derived in the discrete-time formulation as Markov chains \cite{pmlr-v37-sohl-dickstein15, ho2020denoising} before the introduction of continuous-time diffusion models building upon stochastic and ordinary differential equations \cite{song2020score, karras2022elucidating}. As generative models, it has been observed that denoising diffusion models produce high-quality samples both in unconditional \cite{karras2022elucidating, song2020score} and conditional image generation \cite{kawar2022denoising, dhariwal2021diffusion, saharia2022photorealistic, Rombach2021} and outperform previous approaches, notably \textit{Generative Adversarial Networks} (GANs) \cite{dhariwal2021diffusion} as they do not exhibit mode collapse. Diffusion models proved their synthesis capabilities not only for images but also for video \cite{blattmann2023videoldm, ho2022video}, audio \cite{kong2020diffwave, rouard2021crash, Koizumi2022}, text \cite{austin2021structured, li2022diffusion}, and protein design \cite{yim2023se, wu2024protein}.
In addition to the synthesis performances, diffusion models are now used for a variety of tasks, from image segmentation \cite{amit2021segdiff, khani2023slime} to inverse problems solving \cite{kawar2022enhancing}.

In remote sensing, the application of diffusion models to Earth Observation is relatively new. RSDiff~\cite{sebaq2023rsdiff} trains a cascade of a low-resolution diffusion model that feeds into a text-conditioned super-resolution diffusion model that synthesizes remote sensing images based on textual descriptions. SatDM \cite{baghirli2023satdm} uses a conditional DDPM, conditioned on building footprints, to generate new labeled samples to train deep networks for building extractions. \citeauthor{espinosa_generate_2023} \cite{espinosa_generate_2023} extends this idea to a DDPM conditioned on semantic maps using ControlNet~\cite{Zhang_2023_ICCV} to generate new synthetic aerial imagery of Scotland.
While generating fake remote sensing imagery has limited practical applications, diffusion models have been adapted to solve classical EO tasks, such as change detection and cloud removal \cite{Sanguigni2023DiffusionMF} and super-resolution \cite{rs14194834, rs15133452, xiao2023ediffsr}.
In this work, we will leverage the modeling ability of diffusion models not to generate new data, but to estimate the \emph{plausibility} of EO images and identify unusual observations, \ie find images that are out of the distribution of ``usual'' acquisitions.

\subsection{Out-of-distribution detection}

\textit{Out-of-distribution} (OOD) detection is a special case of anomaly detection. Early works focused on detecting OOD samples that fell outside the training dataset of machine learning models, to prevent failure in the predictions.
The first approaches required a dataset of ``unusual'' samples in order to train OOD detectors in a supervised manner \cite{lee2018training, malinin2018predictive}. These approaches restrict the definition of the OOD samples since we are not considering all possible OOD outside the distribution of interest \cite{charpentier2020posterior, wang2022vim}. Recent approaches are \textit{post-hoc}, \ie they rely on features extracted from pre-trained models. In this way, we avoid retraining neural networks for the specific task of OOD detection and benefit from existing pretrained features. 
In general, the goal is to assign a \emph{score} to an observation. If it is greater than a threshold, then the image is considered in-distribution, and OOD otherwise.

Numerous OOD scorers assume that there is a discriminative model, \eg a classifier or a segmenter, that has been trained on some dataset.  \textit{Post-hoc} scorers try to discriminate between out-of-distribution and in-distribution samples based only on the predictions of a classifier or segmentation model. They assess the model certainty on its prediction and samples whose predictions have high uncertainty scores are considered OOD \cite{liu2020energy, macedo2021entropic}. However, there has recently been a surge in fully unsupervised OOD scorers, \ie scorers that do not depend on a discriminative model. These scorers can therefore be used even on unlabeled datasets. For example, some post-hoc scorers have been introduced to work on the features learnt by self-supervised models \cite{sehwag_ssd_2020,mohseni_selfsupervised_2020}.

Similarly, reconstruction-based OOD scorers rely on generative models to reproduce an image given a corrupted version of it. The corruption can be \eg downsampling for \textit{Autoencoders} (AE) and \textit{Variational Autoencoders} (VAE) \cite{DBLP:journals/corr/abs-1812-02765, 9878470, zong2018deep} or noise for diffusion models \cite{Graham_2023_CVPR, Gao_2023_ICCV, 10.5555/3618408.3619344}. In principle, the generative model has learned the training distribution and should have lower reconstruction errors on in-distribution samples (ID) than on OOD samples. The reconstruction error can be evaluated with different similarity metrics depending on the nature of the samples. For images, common metrics are the Mean-Squared Error (MSE) and the Learned Perceptual Image Patch Similarity (LPIPS), the latter being more aligned with human perception \cite{zhang2018unreasonable}.
As diffusion models proved robust reconstruction faculty, several OOD scorers employ them in the context of images denoising \cite{Graham_2023_CVPR, Gao_2023_ICCV} or inpainting \cite{10.5555/3618408.3619344}. In addition, as diffusion models have a variational interpretation \cite{kingma2021variational}, \citet{goodier2023likelihood} proposed an OOD scorer based on diffusion models' ELBO, linking the reconstruction error to the statistical likelihood.

In remote sensing, OOD detection has been a somewhat niche topic.
To the best of our knowledge, the only works on detecting OOD remote sensing images are those from \citet{gawlikowski2021out} and \citet{coca_hybrid_2022}. \citet{gawlikowski2021out} introduced an OOD scorer based on a Dirichlet Prior Network linked to a classification model trained on remote sensing images to detect a shift in the classes, in sensor characteristics, or in geographical areas that would reduce classification accuracy. \citet{coca_hybrid_2022} later adapted this work to a self-supervised model to detect satellite images containing burned areas. As these images were rare in the training dataset, they were able to characterize them as OOD. We will show that this approach is a promising avenue to detect unusual events in remote sensing imagery.

\begin{figure*}[t]
  \centering
   \includegraphics[width=\linewidth]{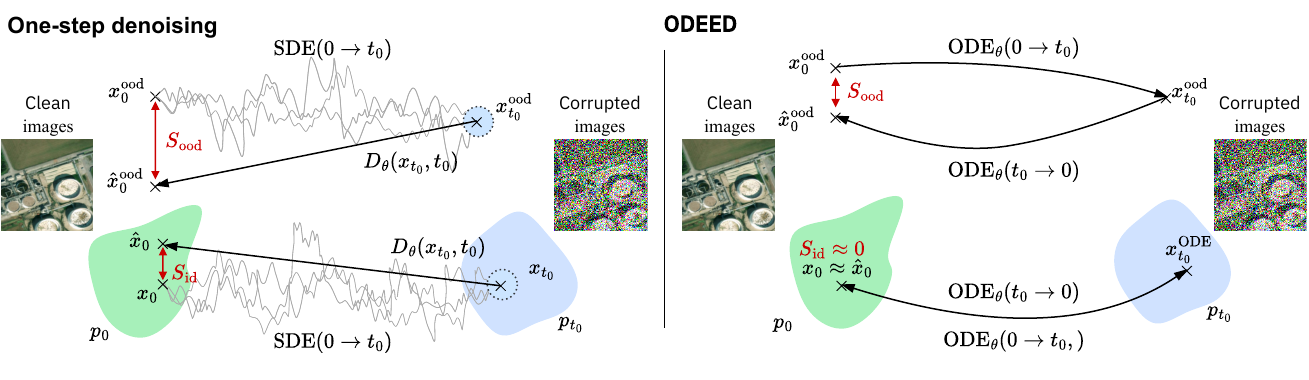}
  \caption{\textbf{Illustration of the one-step denoising and \ours scorers.} (Left) The one-step denoiser samples multiple corrupted versions of the original image $x_{t_0} \sim p_{t_0}(x | x_0)$ thanks to the forward SDE and then evaluates similarity scores on the one-step reconstructions made with the diffusion model $D_\theta$. (Right) The \ours scorer encodes the initial image into a unique latent $x_{t_0}$ with the PF-ODE estimated with $D_\theta$ and then decodes the latent. For the true PF-ODE, in-distribution samples' reconstruction is perfect.}
  \label{fig:method}
\end{figure*}

\section{Methods}
\label{methods}

In this work, we investigate diffusion models for OOD detection and focus on three diffusion-based scorers: \begin{itemize}
    \item \textbf{Diffusion loss scorer} based on the time-truncated diffusion losses which average reconstruction errors;
    \item \textbf{One-step denoising scorer} focusing on the denoising performances at fixed timestep;
    \item \textbf{\ours (ODE Encoding Decoding) scorer}, a new OOD scorer leveraging PF-ODE trajectory accuracy as a way to discriminate between in and out-of-distribution samples. 
\end{itemize}

\subsection{Background on Diffusion Models}

Continuous-time diffusion models generalize discrete-time diffusion models to the infinite timesteps case. They do not rely on Markov chains but on \textit{Stochastic Differential Equations} (SDEs) instead, both for the diffusion and the denoising processes. 
Let us consider a diffusion process $\{x_t\}_{t \in [0, T]}$, with fixed boundary conditions $x_0 \sim p_0(x) = p_\text{data}(x)$ and $x_T \sim p_T(x) = p_\text{prior}(x)$, and $p_t(x_t)$ denote the marginal density at time $t$. This diffusion process is a solution to the following SDE:
\begin{equation}
\label{forwardSDE}
dx_t = f(x_t, t)dt + g(t)dw_t
\end{equation}
where $f: \mathbb{R}^d \times [0, T] \rightarrow \mathbb{R}^d$ is the \textit{drift}, $g: [0, T] \rightarrow \mathbb{R^*_+}$ is the \textit{diffusion} coefficient and  $w_t$ is the standard Wiener process (\ie Brownian motion). This diffusion process corrupts the original image $x_0$ into a Gaussian noise $x_T$.
We know from \cite{ANDERSON1982313} that the forward SDE  in \cref{forwardSDE} admits a reverse SDE mapping $p_T$ to $p_0$: 

\begin{equation}
\label{reverseSDE}
dx_t = \left[f(x_t, t) - g(t)^2\nabla_x\log p_t(x_t) \right]dt + g(t)d\Bar{w_t}
\end{equation}
Moreover, these diffusion processes share the property of admitting a \textit{deterministic} process preserving the same marginal distributions $p_t$ for all $t \in [0, T]$. It is described by the \textit{Probability Flow} Ordinary Differential Equation (PF-ODE):
\begin{equation}
\label{ode}    
dx_t = \left[f(x_t, t) - \frac{g(t)^2}{2}\nabla_x \log p_t(x_t)\right]dt
\end{equation}
Then, the ODE allows encoding samples from the data distribution into the prior distribution by integrating \cref{ode} from $t=0$ to $t=T$ and vice-versa.

Both the backward SDE \cref{reverseSDE} and the PF-ODE \cref{ode} equations include the \emph{score function} $\nabla_x\log p_t$ that has no known close form in practical cases. In this paper, we are embracing the diffusion framework proposed by \cite{karras2022elucidating}, in which the score function derives from a perfect denoiser $D$:

\begin{equation}
\label{score_approx}
    \nabla_x \log p_t(x) = \frac{1}{t^2}\Big(D(x, t) - x\Big)
\end{equation}

We approximate this denoiser with a neural network $D_\theta$ optimized to reduce the reconstruction loss at every time $t$: 

\begin{equation}
\label{loss}
    \mathbb{E}_{x_0\sim p_0}\mathbb{E}_{\epsilon \sim \mathcal{N}(0, t^2I)}\left[ \lambda(t) \lVert D_\theta(x_0 + \epsilon) - x_0 \rVert_2^2\right]
\end{equation}

\subsection{Detecting OOD with diffusion models}

In addition to the generation abilities of continuous-time diffusion models, they can be used in the context of out-of-distribution detection. We leverage three scorers using diffusion models, based on their reconstruction performances.

 First, we want to evaluate the effectiveness of the loss function as an OOD scorer. The intuition behind this is that diffusion models should generalize to in-distribution samples unseen during train but fail on OOD samples, resulting in a greater loss. Since in our case, both in and out-of-distribution are satellite images, we argue that the differences in reconstruction can be observed at small times. Indeed, diffusion models generate images by first placing the overall structure (large times) and then fill-in the details (small times) \cite{biroli2024dynamical, raya2024spontaneous}. Thus, we compute the reconstruction error \cref{loss} for several times below a given threshold $t_0$. In other words, for a given image, we sample $N$ noise corrupted versions of it with increasing times $t_1, \dots, t_N$, then we estimate the $N$ reconstructions with the learned diffusion model $D_\theta$ and compute the weighted average loss:
\begin{equation}
\label{scorer_loss}
S^{\text{loss}}_{t_0}(x_0) = -  \mathbb{E}_{t, \epsilon}\left[\lambda(t)\lVert D_\theta(x_0 + \epsilon) - x_0 \rVert_2^2\right]
\end{equation}
where $t \sim \mathcal{U}(0, t_0)$ and $\epsilon \sim \mathcal{N}(0, t_0^2I)$.
We experiment with two different weighting functions $\lambda(t)$: the training weighting function \cite{karras2022elucidating} and a linear weighting as in \cite{goodier2023likelihood}.

Second, as the learned diffusion model $D_\theta$ is a denoiser conditioned on time, we leverage its reconstruction performances in \textit{one-step denoising}, as shown in \cref{fig:method} (left). For several values to $t_0$, we investigate the ability of the model to reconstruct an initial image $x_0$ from a noised version $x_t \sim p_t(x_t | x_0)$. To do so, we sample several noises and average a reconstruction error across the samples, considering a single denoising step:
\begin{equation}
\label{score_denoiser}
    S_{t_0}^{\text{denoiser}}(x_0) = - \mathbb{E}_{\epsilon \sim \mathcal{N}(0, t_0^2I)}\big[d( D_\theta(x_0 + \epsilon) - x_0)\big]
\end{equation}
where $d(\cdot, \cdot)$ is a similarity measure, \eg MSE or LPIPS.
In practice, we observe that the denoiser $D_\theta$ reconstructs well from noised versions of $x_0$ for small values of $t_0$.

\subsection{\ours scorer for OOD detection}
\label{sdevsode}

Due to the stochastic nature of the diffusion process using the forward SDE \cref{forwardSDE}, proceeding to several integrations of the forward SDE from an initial image $x_0$ to time $t_0$ produces multiple $x_{t_0} \sim p_{t_0}(x_{t_0}|x_0)$. It is also true that reversing the diffusion process from a latent $x_T$ with the backward SDE \cref{reverseSDE} generates several denoised images. This means that stochastic samplers lead to various reconstructions that can be more or less distant from the initial image, depending on the level of stochasticity used, introducing estimation errors in the reconstruction scores.

In contrast, integrating the PF-ODE \cref{ode} from $0$ to $t_0$ encodes the initial image $x_0$ into a \textit{unique} noisy latent $x_{t_0}$. Conversely, integrating it from $t_0$ to $0$ allows us to get back to the exact initial image. In other words, the trajectory is fully deterministic. This property is fundamental in the context of reconstruction-based scorers, as diffusion models can propose exact reconstructions of the initial images.

Based on this observation, we introduce \textbf{\ours} a novel reconstruction-based OOD scorer based on the PF-ODE trajectories (\cref{fig:method} right). To do so, we approximate the true score function in \cref{ode} with the diffusion model $D_\theta$ following \cref{score_approx}.
Let $\text{ODE}_\theta(\cdot, t_1 \rightarrow t_2)$ denote the operation of integrating the $D_\theta$-approximated ODE from $t_1$ to $t_2$ with a numerical solver. Our scorer operates in two phases:

\begin{enumerate}
    \item  \textit{Encoding}: we encode the clean image $x_0$ into a unique noisy latent $x_{t_0}^\text{ODE}$ by integrating the ODE from the starting time $t=0$ to the intermediate time $t=t_0$.
\begin{equation}
    x^\text{ODE}_{t_0} = \text{ODE}_\theta(x_0, 0\rightarrow t_0)
\end{equation}

    \item \textit{Decoding}: we solve the ODE with decreasing times from $t_0$ to $0$, \ie denoising the latent to a clean image that is the reconstruction $\hat{x}_0^\text{ODE}$ of the original sample.

\begin{equation}
    \hat{x}_0^\text{ODE} = \text{ODE}_\theta(x_{t_0}^\text{ODE}, t_0\rightarrow 0)
\end{equation}
\end{enumerate}

Then, we define the OOD scorer from the similarity between the reconstructed and the original images.
\begin{equation}
     S_{t_0}^\text{ODE}(x_0) = -d(x_0, \hat{x}_0^\text{ODE}) 
\end{equation}

Once again, this difference measures how well the diffusion model can reconstruct the clean observation $x_0$. Compared to previous scorers, it is entirely \textit{deterministic} and therefore should better represent whether $x_0$ belongs to the modeled distribution, without being influenced by the stochastic trajectories.

\section{Experiments}
\label{experiments}

\begin{figure}[t]
    \begin{minipage}{0.5\textwidth}
        \centering
        \includegraphics[width=0.8\linewidth]{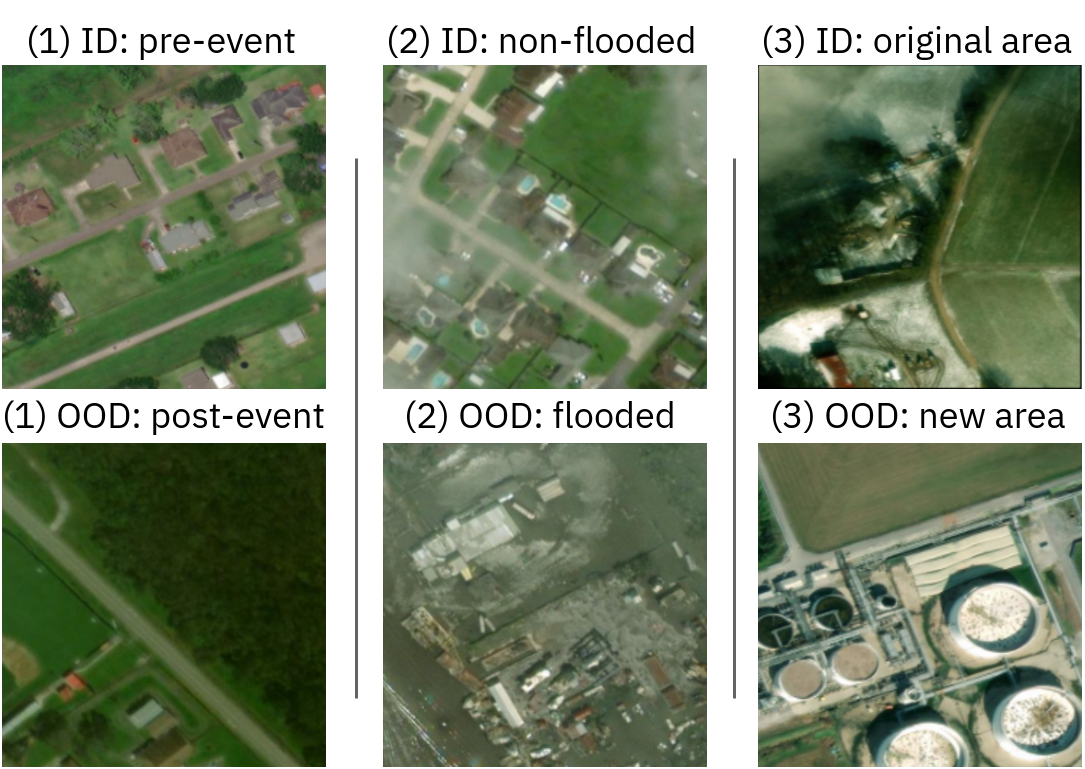}
        \caption{The three scenarios derived from the  \SN dataset. The first two setups focus on the impact of floodings while the third one centers around geographical domain shift.}
        \label{fig:spacenet}
    \end{minipage}
\end{figure}

\subsection{Toy Problem: Cloud Cover Detection}

To test the relevance of our approaches, we experiment with the simple task of discriminating between cloud-free images and cloudy images.
Cloud detection can be achieved with well-designed features and deep learning techniques. Nevertheless, it is an easy use case on which we want to test the reconstruction-based approaches with diffusion models.

\paragraph{Experimental setup}
For this task, we employ the pre-trained discrete-time diffusion model from \cite{Sanguigni2023DiffusionMF}, trained on 8000 $64\times64$ pixels RGB cloud-free images from the Sentinel-2 Cloud Mask Catalogue dataset \cite{francis_2020_4172871}.
Cloud-free images are in-distribution samples and images with cloud coverage above 10\% are OOD. Since the model was only trained on cloud-free images, we expect it to fail to reconstruct images with clouds. As the model is discrete time, we first consider only the one-step denoising scorers. We also use the mean pixel intensity as a naive scorer baseline.

\begin{table}[t]
  \centering
  \begin{tabular}{llcl}
 Method& AUC \textuparrow& FPR95\% \textdownarrow &$R^2$ \textuparrow\\
 \midrule
 Image mean& \textbf{88.3}&41.6 &\textbf{58.4}\\
 Diffusion MSE & \underline{85.1}& \textbf{29.1}&\underline{49.4}\\
 Diffusion LPIPS & 84.3& \underline{29.2}&37.6\\
 \end{tabular}
  \caption{Results for OOD detection on the Sentinel-2 Cloud Mask Catalogue where OOD samples are images with cloud coverage above 10\%. Diffusion models-based approaches are competitive with the baseline and show good results for both AUC and FPR95\%}
  \label{tab:toy_results}
\end{table}

\subsection{\SN}

We then extend our experiments to more challenging applications by considering the \SN dataset \cite{hansch_spacenet_2022}.

\paragraph{Dataset}
\SN is a dataset of paired pre- and post-flooding images from East Louisiana and Germany. It is composed of 3-band RGB images from Maxar satellites. It includes segmentation masks for both the roads and the buildings while distinguishing between flooded and non-flooded objects. We downsample the original images by a factor of 2 to produce $256\times256$ patches and apply overlap on the training set. We obtain 5864 pre and post-event pairs for the Germany subset and 17660 pairs for Louisiana.

\paragraph{Experimental setup}
Using the images and annotations, we derive three different scenarios of OOD detection on \SN, each using a different definition of OOD. This allows us to test the abilities of different OOD scorers to detect various types of ``out-of-distribution'', shown in \cref{fig:spacenet}: 

\begin{enumerate}
    \item \textbf{Pre-flood vs Post-flood images}:
    in this scenario, the OOD detector aims at discriminating between images prior to the flood (in-distribution samples) and images after the flood (out-of-distribution) within the same geographical domain 
    (Germany or East-Louisiana).   
    
    \item \textbf{Flooded vs non-flooded areas images}: in this scenario, the OOD detector should discriminate between images with no flood (in-distribution) and images with a visible flood (out-of-distribution). In-distribution images include post-event images that do not have a visible flood. The labeling is done based on the segmentation masks: an image is considered OOD if it contains pixels belonging to the ``flooded building'' or ``flooded road'' class in the ground truth.\footnote{Note that this proxy is perfectible and has some failure cases (\ie flooded areas with no roads or buildings are considered ``non-flooded''), which might inflate the number of false positives. FPR results for this setup should therefore be interpreted as a higher bound.}
    
    \item \textbf{Domain OOD}: in this case, we want to detect a geographical shift, \ie discriminating images taken in East Louisiana from images taken over Germany. We consider only pre-event images in this setting.
\end{enumerate}

We want to highlight that in all scenarios, the diffusion models are trained on \textit{pre-event images only} and without any annotations, nor post-event images.

\begin{table*}[!t]
  \centering
\resizebox{\textwidth}{!}{%
  \begin{tabular}{lllcrlll}
  && \multicolumn{2}{c}{\textbf{Pre-flood/Post-flood}} & \multicolumn{2}{c}{\textbf{Non-flooded/flooded}}& \multicolumn{2}{c}{\textbf{Domain OOD}} \\
 & Method & Germany& Louisiana& Germany& Louisiana& Germany & Louisiana\\
 \midrule
 \multirow{4}{*}{\parbox{2cm}{\textbf{Discriminative}\\ Segmentation}}& MPC & 52.6/98.9& 41.1/98.4& 67.4/80.0& 61.0/94.3& 69.8/93.4&47.7/95.5\\
 & Neg-Entropy & 59.6/97.7& 42.8/95.9& 64.3/80/0& 60.7/97.1& 76.3/92.2&48.8/94.3\\
 & DeepKNN (k=5)& 51.9/80.7& 76.4/38.9& 70.1/86.7& 54.3/100& \textbf{93.6}/\underline{28.7}&\underline{80.9}/\textbf{38.6}\\
 & Energy Logits& 67.9/88.6& 67.9/88.6& 56.4/80.0& 60.9/97.1& 84.7/70.9&50.4/92.0\\
 \midrule
 \multirow{2}{*}{\parbox{3cm}{\textbf{Generative}\\ Diffusion Loss}}& Training& 50.4/84.1& 52.3/88.5& 59.7/100& 71.0/71.4& 67.0/86.0&55.9/98.8\\
 & Linear& 49.2/85.2& 53.2/88.5& 70.0/100& 75.7/71.4& 60.4/83.1&57.0/98.0\\
 \midrule
 \textbf{Reconstruction based} && &&&&&\\
\textbf{Autoencoder}&MSE& 28.4/96.6& 26.3/95.1& 57.9/100& 68.5/85.7& 84.9/79.9&  28.3/96.6\\
 &LPIPS& 21.1/97.7& 27.5/96.3& 55.6/100& 69.5/77.1& 75.2/54.9&41.4/100\\
 &Mahalanobis& 48.9/95.1& 30.8/97.1& 51.4/94.9& 72.5/90.6& 49.6/95.0&52.0/94.8\\
 \textbf{1-step denoising} &MSE & 28.8/81.8& 42,2/73.3& 68.5/100& 73.4/77.1& \underline{86.3}/\textbf{20.9}&  33.5/97.7\\
 &LPIPS & \underline{74.5}/\underline{59.1}& \underline{90.9}/\underline{35.7}& 60.5/100& \underline{76.8}/\underline{65.7}& 79.6/52.5&\textbf{82.1}/85.2\\
 \textbf{\ours} (Ours)&MSE & 65.6/76.1& 69.0/80.3& \textbf{83.6}/\textbf{33.3}& \textbf{86.9}/\textbf{42.9}& 41.2/97.1&  60.9/95.4\\
 &LPIPS & \textbf{87.9}/\textbf{20.5}& \textbf{94.5}/\textbf{24.6}& \underline{75.3}/\underline{73.3}& 64.1/85.7& 54.3/97.5&68.3/\underline{70.4}\\
 \bottomrule
 \end{tabular}}
  \caption{Results for OOD detection on the Germany and Louisiana subsets of \SN (AUC $\uparrow$ / FPR95\% $\downarrow$). \ours in combination with the LPIPS metric yields the strongest results on the pre/post flood scenario for both domains. Paired with MSE, \ours delivers the highest AUC and lowest FPR95\% for the non-flooded/flooded scenario. }
  \label{tab:main_results}
\end{table*}

\textbf{Baselines.}
We compare the diffusion models and \ours with several well-established OOD detection techniques.

As a reconstruction-based baseline, we train an autoencoder on pre-event images for each geographical domain.
We use either the MSE or the LPIPS distance on the reconstruction in image space as a score. We also consider the Mahalanobis distance $D_M$ in the latent space \cref{mahalanobis} and estimate the covariance matrix $\Sigma$ and the mean $\mu$ on the train set \cite{DBLP:journals/corr/abs-1812-02765, bandara2022ddpm}. The final score is then a linear combination of the latent space and image space distances:

\begin{align}
\label{mahalanobis}
    S^\text{M} =& - \Big[ \alpha D_M(E(x)) + \beta \lVert x - \hat{x} \rVert_2 \Big]\\
    \text{where } D_M(z) =& \sqrt{(z- \mu)\Sigma^{-1}(z-\mu)} 
\end{align}

For exhaustiveness, we also compare our models to OOD scorers-based discriminative models, even though these models require to be trained on an annotated dataset in the first place. We train a segmentation model $f_\theta$ on \SN for each pre-event domain to predict the classes ``background'', ``roads'' and ``buildings''.
We then extract OOD scores from this model using standard approaches from the literature:
\begin{itemize}
    \item Maximum Class Probability (MCP) \cite{jiang2018trust, nguyen2015deep} simply consider that if the model is confident about a prediction then the maximum predicted class probability will be high.
\begin{equation}
    S^\text{MCP}(x)= \max_k P(k = y | x) = \max_k f_\theta(x)
\end{equation}
    \item Negative Entropy: if the model is  unconfident about its prediction then the probability mass will be spread across all classes, resulting in a high entropy: 
\begin{equation}
    S^\text{neg-H}(x) = -H(p) = \sum_{k=0}^{K-1} P(k=y|x) \log P(k=y | x)     
\end{equation}
    \item Energy-Logits (EL) \cite{liu2020energy} computes the \textit{free-energy} on the logits distribution. We denote $g_\theta$ the sub-network of $f_\theta$ predicting the logits.  
\begin{equation}
   S^\text{EL}(x) = -E(x; g_\theta) = T \log \sum_{k=0}^{K-1}e^{g_\theta^{(k)}(x)/T} 
\end{equation}

\item DeepKNN \cite{sun2022out}, which applies a k-nearest neighbors algorithm to the features extracted from the last layer.

\end{itemize}

\paragraph{Diffusion models}
Our diffusion models are trained based on \citet{karras2022elucidating}, 
using the unconditional U-Net backbone from \cite{song2020score} with [64, 128, 128, 256, 256] channels in the downsampling blocks and the reverse for the upsampling blocks.
Time information is sinusoidally embedded on 64 channels and then fed to a linear layer and SiLU activation. Models are trained for 40000 steps with a batch size of 80 (approximately 48 hours), a learning rate of $2e^{-4}$, and AdamW optimizer. To improve training stability, we use gradient clipping and exponential moving average.
We sample using the second-order Heun solver \cite{karras2022elucidating} with $L=20$ discretization steps except in the case of small noise levels ($t_0<1$), where we integrate with 5 steps, as too many sampling steps hinder sampling quality.
$t_0$ hyperparameter is tuned on a held-out validation set.

\paragraph{Autoencoders}
For a fair comparison, we derive the AE architecture from the U-Net backbone of the diffusion model by
removing the time embedders and skip connections.
We train the autoencoders for 20000 steps with a batch size of 80, a learning rate of $2e^{-4}$, and AdamW optimizer.
We set $h=512$ for the latent space dimension as it achieved the best reconstruction results on the validation set. Coefficients $\alpha$ and $\beta$ of the Mahalanobis distance \cref{mahalanobis} are tuned on the validation set to match the reciprocal of the standard deviation of their respective distance \cite{DBLP:journals/corr/abs-1812-02765}.

\paragraph{Segmentation models}
We train DeepLabV3 segmentation models \cite{chen2017rethinking} for a maximum of 1400 steps with a batch size of 32 and use early stopping. We set the learning rate to $2e^{-4}$ and optimize the model's parameters with AdamW.

\section{Results}

We evaluate our OOD detectors using two standard binary classification metrics, by varying the detection threshold: AUC (Area Under Curve) and FPR95\% (False Positive Rate at Recall 0.95). For the latter, note that ID samples are considered as positives and OOD samples are negatives, \ie FPR95\% denote how many OOD samples are wrongly considered as in-distribution for a 95\% recall of the ID samples.

\subsection{Preliminary results: cloud cover}

We report in \cref{tab:toy_results} OOD detection results on the cloud cover dataset. We observe that the diffusion model's reconstruction error is indeed higher on cloudy images, as shown by a high AUC using the diffusion + MSE score. This was expected as the model is trained only on cloud-free images. Diffusion-based methods produce competitive results in terms of AUC with respect to the naive baseline and achieve the best results in terms of FPR95\% with a gain of 12.5 points for the diffusion MSE score against the naive baseline.

To evaluate the calibration of the scores, we perform a linear regression of the OOD score with the cloud coverage percentage. The naive baseline (mean pixel intensity) achieves the highest $R^2$ coefficient (more clouds implies more bright pixels). However, diffusion model scorers also correlate well with the cloud coverage, indicating that more cloudy images tend to be classified as ``more OOD''.
These first results show that diffusion-based scorers discriminate between in and out-of-distribution samples on the simple task of cloudy image detection. This suggests their effectiveness for use cases where no naive baselines exist.

\subsection{Main results}

\begin{figure}[t]
\vspace{12pt}
\begin{minipage}[t]{\linewidth}
\vspace{0pt}
\begin{minipage}[t]{0.49\linewidth}
    \vspace{0pt}
        \centering
    \includegraphics[width=1\linewidth]{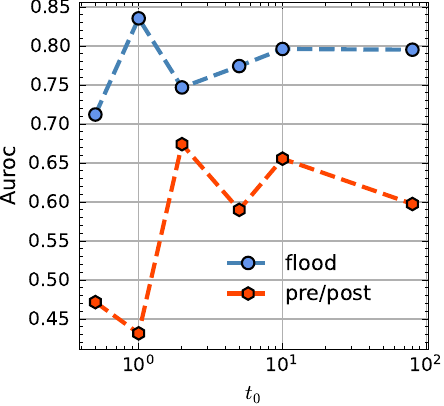}
    \subcaption{$t_0$ \vs {\small AUC}$\uparrow$}
    \label{fig:flood_pre_post_germany_sode_mse}
\end{minipage}
\begin{minipage}[t]{0.49\linewidth}
\vspace{0pt}
        \centering
    \includegraphics[width=1\linewidth]{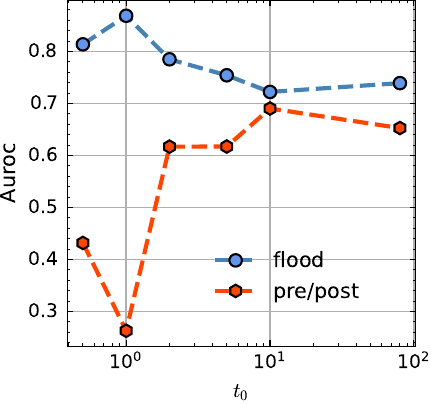}
    \subcaption{$t_0$ \vs {\small AUC}$\uparrow$}
   \label{fig:flood_pre_post_louisiana_sode_mse}
\end{minipage}
\caption{Impact of $t_0$ on AUC for \ours + MSE metric on the geographical domains (a) Germany and (b) Louisiana.}
\label{fig:flood_pre_post_sode_mse}
\end{minipage}
\end{figure}

\begin{figure}[t]
\vspace{12pt}
\begin{minipage}[t]{\linewidth}
\vspace{0pt}
\begin{minipage}[t]{0.49\linewidth}
    \vspace{0pt}
        \centering
    \includegraphics[width=1\linewidth]{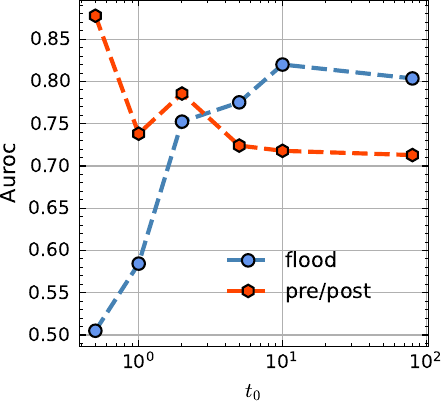}
    \subcaption{$t_0$ \vs {\small AUC}$\uparrow$}
    \label{fig:flood_pre_post_germany_sode_lpips}
\end{minipage}
\begin{minipage}[t]{0.49\linewidth}
\vspace{0pt}
        \centering
    \includegraphics[width=1\linewidth]{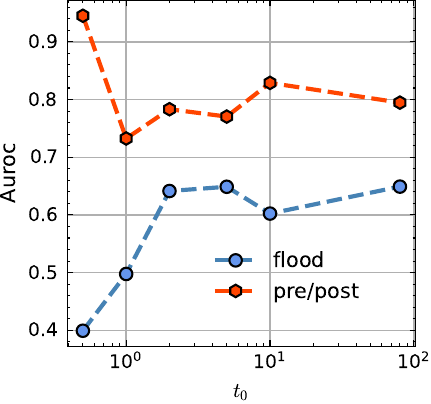}
    \subcaption{$t_0$ \vs {\small AUC}$\uparrow$}
   \label{fig:flood_pre_post_louisiana_sode_lpips}
\end{minipage}
\caption{Impact of $t_0$ on AUC for \ours + LPIPS metric on (a) Germany (b) Louisiana.}
\label{fig:flood_pre_post_sode_lpips}
\end{minipage}
\end{figure}

\begin{table*}[tb]
  \centering
  \setlength\tabcolsep{5pt}
  \begin{tabularx}{\textwidth}{XXcccc|cccc}
  & & \multicolumn{4}{c}{\textbf{Pre-flood/Post-flood}}&\multicolumn{4}{c}{\textbf{Non-flooded/flooded}}\\
  & Domain& \multicolumn{2}{c}{Germany}& \multicolumn{2}{c}{ Louisiana}& \multicolumn{2}{c}{Germany}& \multicolumn{2}{c}{ Louisiana}\\
 & Model& $D_\theta$ G & $D_\theta$ L&   $D_\theta$ L&  $D_\theta$ G&$D_\theta$ G&$D_\theta$ L&$D_\theta$ L& $D_\theta$ G\\
 \midrule
 \multirow{2}{*}{\parbox{3cm}{\textbf{1-step\\ denoising}}} & MSE & 28.8/81.8& 26.2/100&   42,2/73.3&  23.2/98.0&68.5/100&65.6/100&73.4/77.1& 71.9/74.3\\
 & LPIPS & 74.5/59.1& 76.8/51.1&   90.9/35.7&  87.5/31.6&60.5/100&64.5/100&76.8/65.7& 77.4/71.4\\
 \multirow{2}{*}[-0.5em]{\textbf{\ours}} &
 MSE & 65.6/76.1& 61.0/70.5&   69.0/80.3&  80.8/30.7&83.6/33.3&70.0/60.0&86.9/42.9& 76.0/51.4\\
 & LPIPS & 87.9/20.5& 84.0/18.2&   94.5/24.6&  94.6/06.1&75.3/73.3&45.8/80.0&64.1/85.7& 60.8/60.0\\
 \bottomrule
 \end{tabularx}
 
  \caption{Results obtained in OOD detection (AUC $\uparrow$ / FPR95\% $\downarrow$) using, on a given geographical domain, a model trained on another domain. We note that the differences in performance are slight, both in AUC and in FPR95. A minor difference can be seen in the case of ODE encoding in the pre/post scenario, where the high AUC over Louisiana is preserved (94.5 \vs 94.6 when using LPIPS).}
  \label{tab:xdomain}
\end{table*}

We report in \cref{tab:main_results}  the OOD scorer's performances on the Germany and Louisiana subsets of \SN.
In the near-OOD setting (pre-flood/post-flood and non-flooded/flooded), AE exhibits poor OOD detection abilities, with AUC well under the 0.5 random baseline. Discriminative OOD detectors fare better but still misclassify a lot of OOD images as in-distribution, as shown by the high FPR95.
In contrast, the best AUC and FPR95\% results are achieved with our \ours method for the pre/post-event and the non-flooded/flooded scenarios.
On the pre/post-event OOD setup, the combination of \ours + LPIPS outperforms the best baseline by a margin of 13.4 points on the Germany dataset and 3.6 pts on the Louisiana dataset in terms of AUC. We notice a gain of -38.6pts of FPR95\% on the pre/post-event Germany experiment when using the \ours + LPIPS and a gain of -11.1 for the same setup on the Louisiana dataset. 
Interestingly, while best results are obtained with the pair \ours + LPIPS for the pre/post-event setup on both areas, the MSE brings the ODE-encoding to top performances on the non-flooded/flooded scenarios with 83.6 AUC and 33.3 FPR95\% on Germany and 86.9 AUC and 42.9 FPR95\% on Louisiana. We hypothesize that this difference is linked to the difference between ID and OOD samples in the two scenarios: in the first scenario the satellite images have different global aspects, whereas floods are locally spatialized elements of the same global context. 

For all setups, reconstruction-based methods based on diffusion models outperform the autoencoder-based ones. 

The largest performance gap reaches 39.0 (87.9 \vs 48.9) and 63.7 pts (94.5 \vs 30.8) in AUC on the pre/post-flood scenario for respectively Germany and Louisiana. The narrowest gap is set on the Germany/Louisiana domain OOD case where autoencoder-based methods are on par with diffusion models (respectively 85.6 and 86.3 in top AUC).

Yet, the \ours scorer performs poorly on the domain change detection. We suspect that the model generalizes enough to reconstruct images from other domains for small noise levels. This idea is reinforced by the decent performances of the one-step denoisers in this scenario: best AUC (82.1) with LPIPS on the Louisiana dataset and second best on the Germany dataset (86.3) with MSE, while achieving the best FPR95\% (20.9). DeepKNN yields top results in domain OOD on the Germany dataset and second best on the Louisiana domain. This could indicate that segmentation models learn features that are more domain-specific than the reconstruction models, and therefore generalize worse when switching geographical areas.
Finally, \cref{tab:main_results} shows that diffusion model losses, independently of the weighting function used, are poor likelihood estimators in our settings.
The various dynamics of the reconstruction errors may explain the poor OOD detection performance of diffusion losses which average the latter over a large time range.

\paragraph{Effect of $t_0$ on OOD detection}
We now evaluate the impact of the corruption time $t_0$ used in the \ours method, as it can be tuned on a validation set. We observed experimentally that its effect on the OOD detection performances depends on the task and on the similarity measure considered. For example,  \cref{fig:flood_pre_post_sode_mse} highlights that using the MSE as the score results in higher AUC scores for small $t_0$ in the flood OOD scenario, while it reaches the highest performance for medium intermediate times for the pre/post OOD setup. This might indicate that smaller $t_0$ should be preferred for more localised anomalies. On the contrary, \cref{fig:flood_pre_post_sode_lpips} shows that the LPIPS ability to discriminate between the pre and post-event images is higher for early $t_0$ whereas giving the best results on non-flooded/flooded images for intermediate noise levels. We attribute this to the perceptual effect of LPIPS, where localised changes result in small LPIPS variations, and therefore need a higher $t_0$ to isolate flood events.

\begin{figure}[t]
\centering
\begin{minipage}[b]{0.49\linewidth}
  \centering
  \includegraphics[width=\linewidth]{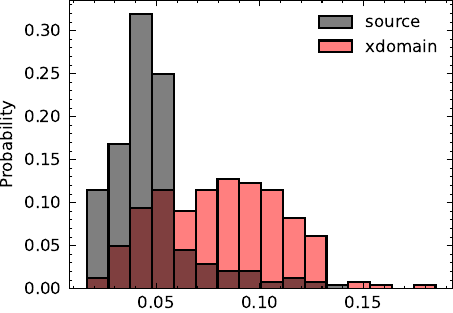}
\end{minipage}
\begin{minipage}[b]{0.49\linewidth}
  \centering
  \includegraphics[width=\linewidth]{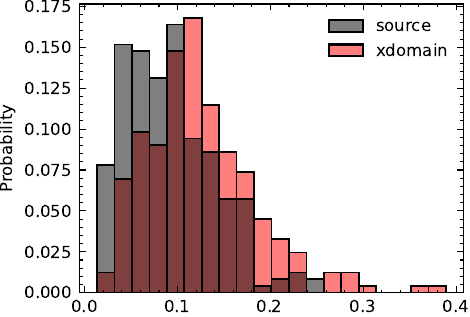}
\end{minipage}
\begin{minipage}[b]{0.49\linewidth}
  \centering
  \includegraphics[width=\linewidth]{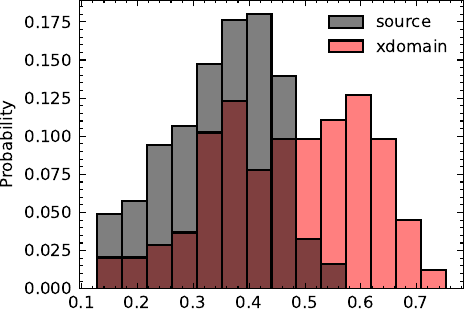}
\end{minipage}
\begin{minipage}[b]{0.49\linewidth}
  \centering
  \includegraphics[width=\linewidth]{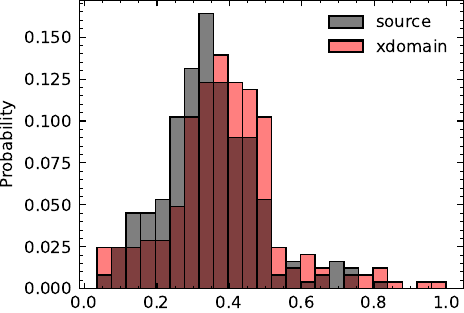}
\end{minipage}
\begin{minipage}[b]{0.49\linewidth}
  \centering
  \includegraphics[width=\linewidth]{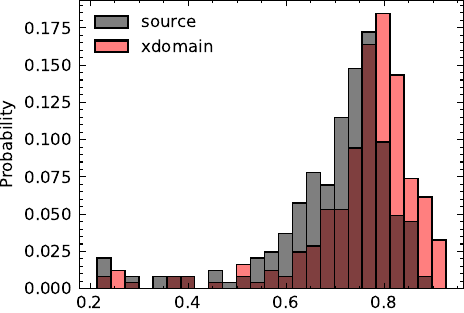}
\end{minipage}
\begin{minipage}[b]{0.49\linewidth}
  \centering
  \includegraphics[width=\linewidth]{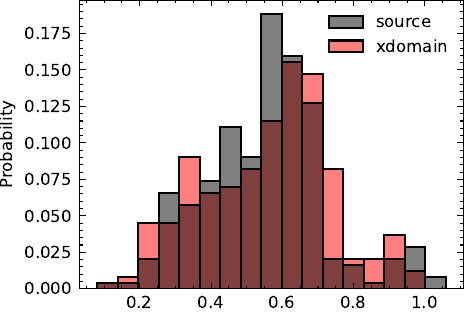}
\end{minipage}
\caption{LPIPS shift on pre/post-event Louisiana OOD in function of $t_0$, from 0.05 (top), 0.5 (middle), to 5.0 (bottom), for ID (left) and OOD (right) samples. On the Louisiana dataset, LPIPS increases when using $D_\theta$ Germany (xdomain) instead of $D_\theta$ Louisiana (source). As $t_0$ increases, the shift is less severe.}
\label{fig:xdomain_shift}
\end{figure}

\paragraph{Cross domain performances}
The poor results of \ours on geographical domain OOD lead us to question the generalization of diffusion models to new areas.  \cref{tab:xdomain} shows that OOD detection performance remains close, even when switching the models' domains. Naturally, reconstruction errors are higher on the domain for which the diffusion model has not been trained, \ie $D_\theta$ Germany provides lower quality reconstruction on Louisiana images than $D_\theta$ Louisiana (and vice-versa). Nevertheless, we show in \cref{fig:xdomain_shift} that the distribution of the scores is shifted both for in and out-of-distribution samples. This means the detection threshold has also to be shifted, however, detection abilities are preserved, as illustrated by the similar AUC and FPR95\% for the same-domain and cross-domain setups. This is promising for the detection of anomalies even under non-stationary data distributions.

\section{Conclusion}
\label{conclusion}

In this work, we evaluated the effectiveness of diffusion models to detect out-of-distribution Earth Observation images. We introduced the \ours scorer that leverages the deterministic reconstruction capabilities of continuous-time diffusion models.
We evaluated these approaches for 1) cloud detection and 2) a challenging ensemble of OOD detection tasks on the Space-Net 8 dataset. We demonstrated that our \ours scorer significantly outperforms baselines in the more challenging flood-related scenarios, showing the interest of diffusion models to detect ``near-out-of-distribution'' remote sensing imagery, such as images of floods. These findings open the door to new approaches to detect rare events from unlabeled EO data using generative modeling.

\small
\paragraph{Acknowledgements}
This research was conducted as part of the research project MAGE (ANR-22-CE23-0010) funded by the \textit{Agence Nationale de la Recherche}.
This work was granted access to the HPC resources of IDRIS under the allocation AD011014327 made by GENCI.

{
    \small
    \bibliographystyle{ieeenat_fullname}
    \bibliography{main}
}


\end{document}